\documentclass{article}

\usepackage[accepted]{icml2018}

\usepackage[utf8]{inputenc} % allow utf-8 input
\usepackage[T1]{fontenc}    % use 8-bit T1 fonts
\usepackage{hyperref}       % hyperlinks
\usepackage{url}            % simple URL typesetting
\usepackage{booktabs}       % professional-quality tables
\usepackage{amsfonts}       % blackboard math symbols
\usepackage{nicefrac}       % compact symbols for 1/2, etc.
\usepackage{microtype}      % microtypography
\usepackage{amsmath,amsfonts}
\usepackage{cleveref}

\usepackage{amsthm}
\usepackage{multirow}
\usepackage{graphicx}
\usepackage{algorithm}
\usepackage{algorithmic}
\usepackage{bm}
\usepackage{bigints}

\newcommand\N{\ensuremath{\mathcal{N}}}

\newcommand{\SPT}{\mathcal{T}}

\newcommand{\SPN}{\mathcal{S}}
\newcommand{\X}{\mathbf{X}}
\newcommand{\F}{\mathbf{F}}

\newcommand{\U}{\mathbf{U}}
\newcommand{\C}{\mathbf{C}}

\newcommand{\K}{\mathbf{K}}
\newcommand{\kk}{\mathbf{k}}
\newcommand{\data}{\mathcal{D}}
\newcommand{\x}{\mathbf{x}}
\newcommand{\y}{\mathbf{y}}
\newcommand{\f}{\mathbf{f}}
\newcommand{\xn}{\mathbf{x}_{n}}
\newcommand{\new}{_{*}}

\newcommand{\ProductNode}{\mathsf{P}}
\newcommand{\SumNode}{\mathsf{S}}
\newcommand{\SplitNode}{\mathsf{G}}

\newcommand{\Leaf}{\mathsf{L}}
\newcommand{\Child}{\mathsf{C}}
\newcommand{\Node}{\mathsf{N}}

\newcommand{\ch}{\ensuremath{\mathbf{ch}}}

\newcommand{\scope}{\ensuremath{\mathbf{sc}}} % leaf function
\newcommand{\w}{w}

\theoremstyle{definition}

\icmltitlerunning{Learning Deep Mixtures of Gaussian Process Experts Using Sum-Product Networks}
\begin{document}
  
  \twocolumn[
\icmltitle{Learning Deep Mixtures of Gaussian Process Experts\\ Using Sum-Product Networks}

\begin{icmlauthorlist}
\icmlauthor{Martin Trapp}{graz,ofai}
\icmlauthor{Robert Peharz}{cbl}
\icmlauthor{Carl E. Rasmussen}{cbl}
\icmlauthor{Franz Pernkopf}{graz}
\end{icmlauthorlist}

\icmlaffiliation{ofai}{Austrian Research Institute for Artificial Intelligence, Vienna, Austria}
\icmlaffiliation{graz}{Signal Processing and Speech Communication Lab., Graz University of Technology, Graz, Austria}
\icmlaffiliation{cbl}{Computational and Biological Learning Lab., University of Cambridge, Cambridge, UK}

\icmlcorrespondingauthor{Martin Trapp}{martin.trapp@ofai.at}
\icmlkeywords{Sum-Product Networks, Gaussian Process, Mixture of Experts, Machine Learning}

\vskip 0.3in
]

\printAffiliationsAndNotice{}  % leave blank if no need to mention equal contribution

\begin{abstract}
While Gaussian processes (GPs) are the method of choice for regression tasks, they also come with practical difficulties, as inference cost scales cubic in time and quadratic in memory.
In this paper, we introduce a natural and expressive way to tackle these problems, by incorporating GPs in sum-product networks (SPNs), a recently proposed tractable probabilistic model allowing exact and efficient inference.
In particular, by using GPs as leaves of an SPN we obtain a novel flexible prior over functions, which implicitly represents an exponentially large mixture of local GPs.
Exact and efficient posterior inference in this model can be done in a natural interplay of the inference mechanisms in GPs and SPNs.
Thereby, each GP is -- similarly as in a mixture of experts approach -- responsible only for a subset of data points, which effectively reduces inference cost in a divide and conquer fashion.
We show that integrating GPs into the SPN framework leads to a promising probabilistic regression model which is: 
(1) computational and memory efficient, 
(2) allows efficient and exact posterior inference, 
(3) is flexible enough to mix different kernel functions, and
(4) naturally accounts for non-stationarities in time series.
In a variate of experiments, we show that the SPN-GP model can learn input dependent parameters and hyper-parameters and is on par with or outperforms the traditional GPs as well as state of the art approximations on real-world data.
\end{abstract}

\section{Introduction}

Due to their non-parametric nature, Gaussian Processes (GPs) \cite{Rasmussen2006} are a powerful and flexible principled way for non-linear probabilistic regression.
In the past years, GPs have had a substantial impact in various research areas, including reinforcement learning \cite{Rasmussen2003}, active learning \cite{Park2011}, preference learning \cite{Chu2005} and parameter optimization \cite{Rana2017}.
However, a limitation of the GP model is that learning scales with $\mathcal{O}(N^3)$ in the number of observations $N$ and has a memory consumption of $\mathcal{O}(N(N+D))$ -- where $D$ is the number of dimensions -- making it unpractical for large datasets.

Several approaches to overcome the computational complexity in GPs have been proposed, which can be categorized into two main strategies.
The first strategy employs sparse approximations of GPs \cite{Williams2000, Candela2005, Hensman2013, Gal2015, Bauer2016}, reducing the inference cost to cubic (time) and quadratic (memory) in the number of so-called inducing points, rather than the number of samples.
By doing so, sparse approximations allow to scale GPs up to reasonably large datasets, comprising millions of samples, while typically using only a few hundred inducing points.

Alternative to sparse approximations, the second strategy aims to distribute the computations by using local models or hierarchies of thereof \cite{Shen2005, Ng2014, Cao2014, Deisenroth2015}.
Those approaches are usually related to the Bayesian Committee Machine (BCM) \cite{Tresp2000} or the Product of Experts (PoE) \cite{Ng2014} approach.
However, as discussed by Deisenroth and Ng \cite{Deisenroth2015} these approaches have either the shortcoming that with an increasing number of local models the predictive variance vanishes or that the posterior mean suffers from weak experts.

In this paper, we propose a natural and expressive framework which falls in the latter strategy of hierarchically composing local GP models.
In particular, we combine GPs with sum-product networks (SPNs), an expressive class of probabilistic models allowing exact and efficient inference.
SPNs represent probability distributions by recursively utilizing factorization (product nodes) and mixtures (sum node) according to an acyclic directed graph (DAG).
The base of this recursion is reached at the leaves of this DAG, which represent user-specified input distributions, each defined only over a sub-scope of the involved random variables.
SPNs typically allow reducing the mechanism of inference to the corresponding mechanisms at the leaves.
For example, marginalization in SPNs is performed by computing the corresponding marginalization tasks at the leaves and evaluating the inner nodes (sums and products) as usual.
Thus, marginalization in SPNs is performed in time linear of the network size, plus the inference costs at the leaves.
So far, SPNs have mainly been applied to density estimation tasks akin to graphical models, typically using Gaussian or categorical distributions as leaves.

However, the crucial insight exploited in this paper is, that SPNs are a sound language for \emph{any} probabilistic model used as leaves.
In particular, it is no problem to use GPs as the leaves of the SPN, immediately yielding an ``SPN over GPs'', or in other words a deep hierarchically structured mixture of local GP experts.
It is easy to see that this model represent a prior over functions, which has, to the best of our knowledge, not been considered before.

We show that in contrast to prior work our approach does not suffer from vanishing predictive variances, allows for \emph{exact posterior inference} and is flexible enough to be used for non-stationary data.
Further, our model is able to mix over different kernel functions weighted by their plausibility, reducing the effect of weak experts.

The rest of the paper is structured as follows.
After reviewing related work and background information on SPNs and GPs,
Section \ref{sec:deepMixtureGP} introduces our SPN-GP model.
In Section \ref{sec:deepMixtureGP}, we further introduce a generic structure learning approach for SPN-GPs, discuss hyper-parameter learning and show to perform exact posterior inference.
We assess the performance of our approach qualitatively and quantitatively in Section~\ref{sec:experiments}.
And finally, discuss and conclude our work in Section~\ref{sec:discussion}.

\section{Related Work}   \label{sec:background}

\subsection{Sum-Product Networks}   \label{sec:spns}
A sum-product network (SPN) \cite{Darwiche2003,Poon2011} over a set of random variables (RVs) $\X$ is a rooted DAG over three types of nodes, namely sums (denoted as $\SumNode$), products (denoted as $\ProductNode$) and leaf distributions (denoted as~$\Leaf$).
A generic node is denoted as $\Node$ and the overall SPN as denoted as $\SPN$.
The leaves are distributions over some subset of random variables $\U \subseteq \X$, pre-specified by the user, where often univariate parametrized distributions (e.g.~Gaussians \cite{Poon2011,Peharz2014a}) are assumed.
This, however, is no requirement, as in fact \emph{any} kind of leaf distributions might be used in the SPN framework.
In particular, we will use GPs as SPN leaves in this paper.

An internal node ($\Node$) computes either a weighted sum or a product, i.e.~%
$\SumNode(\x) = \sum_{\Node \in \ch(\SumNode)} \w_{\SumNode,\Node} \, \Node(\x)$ 
or 
$\ProductNode(\x) = \prod_{\Node \in \ch(\ProductNode)} \Node(\x)$,
where $\ch(\Node)$ are the children of node $\Node$.
Note that each edge $(\SumNode,\Node)$ emanating from a sum node $\SumNode$ has a non-negative weight $\w_{\SumNode,\Node}$, where w.l.o.g.~we assume \cite{Peharz2015} that $\sum_{\Node \in \ch(\SumNode)} \w_{\SumNode,\Node} = 1$.

We require SPNs to be \emph{complete} and \emph{decomposable}, two conditions which are naturally expressed via the \emph{scope} of nodes.
For a leaf $\Leaf$, the scope is defined as the set of random variables $\U$ the leaf is a distribution over, i.e.~$\scope(\Leaf) = \U$.
For an internal node $\Node$ (sum or product), the scope is defined as $\scope(\Node) = \bigcup_{\Node' \in \ch(\Node)} \scope(\Node')$.
An SPN is complete if, for each sum node, all children of the sum node have an identical scope.
An SPN is decomposable if, for each product node, all children have non-overlapping scopes.
These two conditions ensure that sum nodes are proper mixture distributions (with their children as components) and product nodes are factorized distributions (assuming independence among their children).
Thus, SPNs can be seen as a hierarchically structured mixture model, recursively using mixtures (sum nodes) and factorization (product nodes), where the recursion base is reached at the leaf distributions.
One can see that each node in an SPN represents a distribution over its respective scope, but typically the root of the SPN is used as the model.
By this construction, SPNs are both a flexible modelling language and allow exact and efficient inference.
Due to decomposability and completeness, marginalization tasks reduces to the corresponding marginalization tasks at the leaves, over their respective smaller scopes.
Conditioning can be tackled likewise.

Various ways for learning the parameters of SPNs, i.e.~the weights at the sum nodes and the parameters of the leaf distributions, have been proposed.
By interpreting the SPN as a latent variables model, the parameters can be learned using expectation-maximization \cite{Poon2011,Peharz2017}, Bayesian learning \cite{Rashwan2016,Zhao2016,Trapp2016} or a concave-convex procedure \cite{Zhao2016b}. 
Furthermore, Gens and Domingos \cite{Gens2012} trained SPNs discriminatively using gradient descent.
Subsequently, Trapp {\it et al.}~\cite{Trapp2017} introduced a safe semi-supervised learning scheme for discriminative and generative parameter learning, providing guarantees for the performance in the semi-supervised case. 
A major challenge in SPNs is learning a valid structure of such networks. Besides handcrafted approaches \cite{Poon2011,Peharz2014a}, several structure learners \cite{Gens2013,Peharz2013,Rooshenas2014,Vergari2015,Adel2015,Trapp2016,Molina2018} which try to find good structures in a data-driven approach have been proposed.

\subsection{Gaussian Processes}   \label{sec:gps}

A Gaussian process (GP) is defined as any collection of random variables $\F$, where any finite subset of $\F$ is Gaussian distributed, and whereof any two overlapping finite sets are marginally consistent \cite{Rasmussen2006}.
In that way, GPs can naturally be interpreted as distributions over functions $f$, wherein this paper we assume $f \colon \mathbb{R}^D \mapsto \mathbb{R}$.
A GP is uniquely specified by a \emph{mean-function} $m(\x)$ and a \emph{covariance function} $k(\x', \x'')$.
Given pairs of inputs $\xn$ and outputs $f_n$, the joint distribution of $\f = (f_1, \dots, f_N)^T$ is given as a joint Gaussian:
\begin{equation}
 \f \sim \N \left (m(\X), k(\X,\X) \right),
\end{equation}
where the rows of $\X$ are given by $\x_1, \dots, \x_N$ and mean and covariance are given as
\begin{equation}
 m(\X)_i = m(\X_{i,\cdot}) ~~~~~ k(\X,\X)_{i,j} = k(\X_{i,\cdot}, \X_{j,\cdot}).
\end{equation}
For regression, we assume $m\equiv 0$ and $\y = f(\x) + \epsilon$, where $\epsilon \sim \N(0, \sigma_{\epsilon}^2)$ is i.i.d.~Gaussian noise.
Let $\theta = \{\bm \alpha, \sigma_{\epsilon}\}$ denote the parameters of the GP for which $\sigma_{\epsilon}$ is the variance of the noise model and $\bm \alpha$ are kernel function specific hyper-parameters, e.g. in case of a squared exponential kernel function $\bm \alpha = \{\sigma_f, l_1, \dots, l_D\}$ consisting of the variance of the latent function $\sigma_f$ and the respective length-scales $l_1, \dots, l_D$.
Given hyper-parameters, a training set $\data = \{(\xn, \y_n)\}_{n=1}^{N}$ and a test set $\X\new$, our main interest is in the posterior predictive distribution of the corresponding function value.
First, we consider the joint distribution of the training samples and the function values at the test locations $\X\new$, i.e.
\[
   \left[\begin{array}{c}
     \y \\
     \bm f\new
   \end{array} \right]
   \sim \N\left(\bm 0, \left[\begin{array}{cc}
K(\X, \X) + \sigma^2_{\epsilon}\bm I & K(\X,\X\new) \\
K(\X\new,\X) & K(\X\new,\X\new) \end{array} \right]\right) \, .
\]
Therefore, for a single data point $\x\new$ we arrive at the posterior predictive with mean and variance which are given by
\begin{align} 
  \mathbb{E}[f\new] &= \kk_{\new}^T \C^{-1} \y \, ,\label{eq:posteriorGPMean}\\ 
  \mathbb{V}[f\new] &= k(\x\new, \x\new) - \kk_{\new}^T\C^{-1} \kk_{\new} \, ,\label{eq:posteriorGPVar}
\end{align}
where $\C = \K + \sigma^2_{\epsilon}\bm I$, $\K = K(\X, \X)$ and we use $\kk_{\new}$ as a shorthand notation to denote the vector of covariances between the test point and the training data.

As the performance of GPs depend on the selected hyper-parameters, these are typically optimized by maximizing the marginal likelihood, which is given by
\begin{equation} 
  \log p(\y | \X) = -\frac{1}{2} \log \det(\C) + \frac{1}{2}(\y^T \C^{-1} \y) + \frac{N}{2} \log 2\pi \, . \label{eq:mllh}
\end{equation}
The main computational burdens in \Crefrange{eq:posteriorGPMean}{eq:mllh} are the inversion of $\C$ and in \eqref{eq:mllh} additionally the computation of the determinant.
In a naive implementation using Gaussian elimination, all three computations scales in $\mathcal{O}(N^3)$.
These computational burdens together with the memory requirements, which are $\mathcal{O}(N(N+D))$, make GPs unpractical for reasonable large datasets.

\section{Deep Mixtures of Gaussian Processes}   \label{sec:deepMixtureGP}
One common strategy to reduce the inference cost in GP models is to perform the computations by independent experts, where each GP expert is responsible only for a small subset of data points.
First, let us consider the simple case of a single input $x$ and a single output $y$.
A simple yet effective approach to reduce inference complexity is to select a splitting point $t$ and define two local GPs, where one GP is an expert for the data points $x \leq t$ and the other for $x > t$. 
Thus, we have split an infinite collection of RVs into two collections, one indexed by $-\infty \leq t$ and the other by $t < \infty$.
This can also be understood as a fixed gating network \cite{Rasmussen2001}, which assigns all data points $x \leq t$ to one GP expert and all other points to the other GP expert.
Instead of considering a single split point $t$, we can generalize this idea to several split points, each yielding a GP with a different independence assumption.
Finally, by taking a \emph{mixture} over the different splitting positions, we obtain a mixture of GP experts which replaces unconditional independence with a \emph{conditional} independence.

\begin{figure*}%
    \centering
    \includegraphics[width=0.7\linewidth]{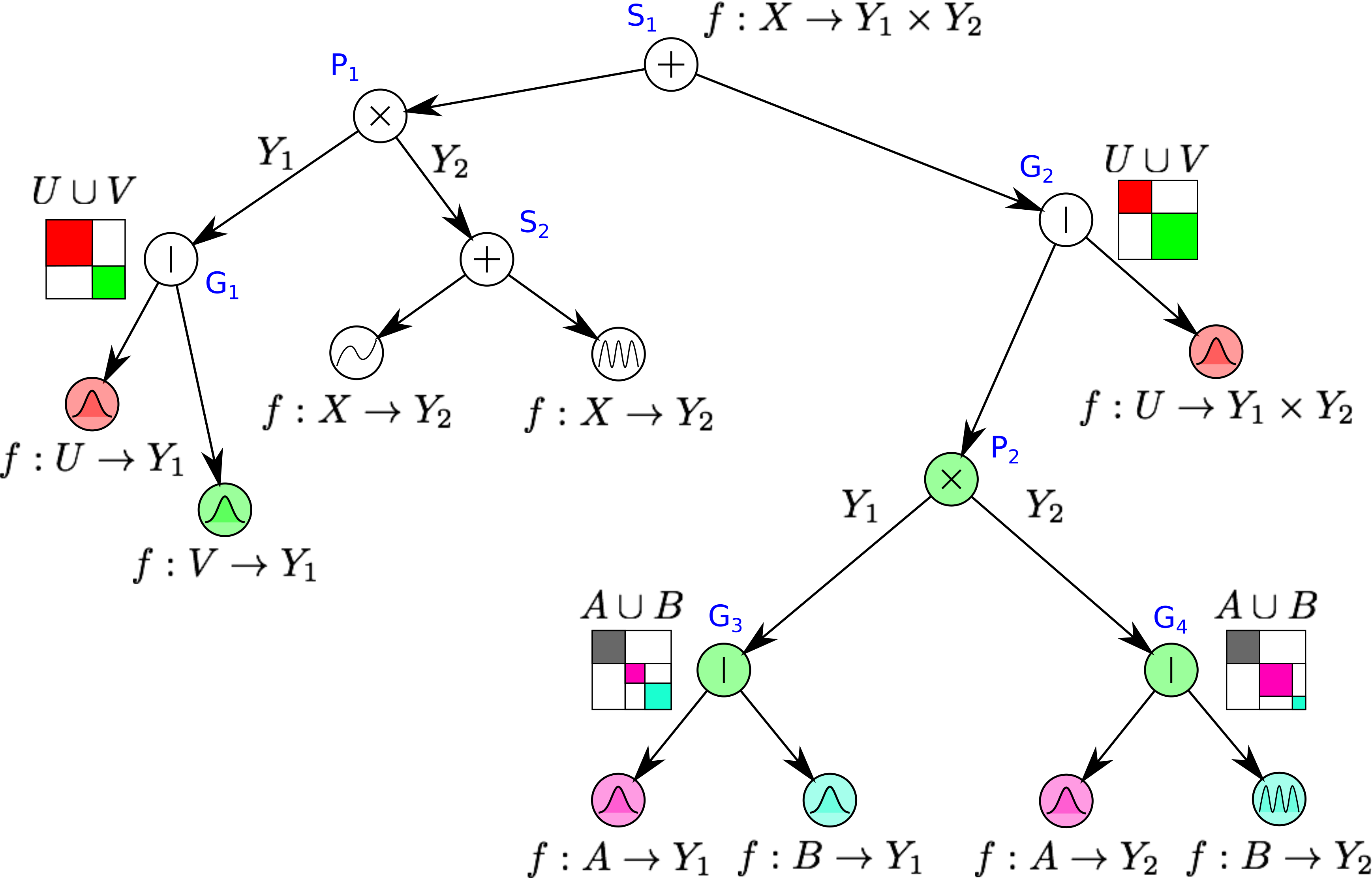}%
    \caption{[Best viewed in colour] Illustration of an SPN-GP model. Sum nodes represent mixtures and product nodes independence assumptions between the output RVs as in classical SPNs. Split nodes (illustrated by a vertical line) represent independence assumptions between the input RVs. Sub-collections of input RVs are colour coded and the symbols at leaf nodes indicate the kernel function of the respective GPs, e.g. squared exponential, polynomial or periodic.}%
    \label{fig:SPN-P}%
\end{figure*}

Incorporating the outlined idea into SPNs results in a natural definition of SPNs with GP leaves.
More precisely, an SPN-GP (SPN with GP leaves) is a hierarchically structured mixture model which recursively combines mixtures over sub-SPN-GPs or GPs with subdivisions of the input space at product nodes.
First, let us consider an SPN consisting of a single sum and $M$ many product nodes, each splitting the infinite collection of input RVs at $C$ different split points into sub-collections.
Further, let all weights of the network be uniform, i.e. $\w_m = \frac{1}{M}$ with $1 \leq m \leq M$.
Therefore, the prior over latent functions $\bm f$ of an SPN-GP is,
\begin{equation}
  p(\bm f | \bm \theta) = \sum_{m=1}^M \w_m \prod_{c=1}^C p(\bm f_{m, c} | \theta_{m, c}) \, ,
\end{equation}
where $\bm f = \{\bm f_{1, 1}, \dots, \bm f_{M,C}\}$ is the set of latent functions of the GP experts and $\bm \theta = \{\theta_{1, 1}, \dots \theta_{M,C}\}$ denotes their respective hyper-parameters.
Clearly, the resulting model is a mixture of GP experts.

More generally, we can express the prior over latent functions of any SPN-GP in terms of a mixture over induced trees \cite{Zhao2016}, i.e.
\begin{equation}
  p(\bm f | \bm \theta) = \sum_{t=1}^\tau \prod_{(\SumNode, \Child) \in \SPT_{t, E}} \w_{\SumNode, \Child} \prod_{\Leaf \in \SPT_{t, V}} p(\bm f_{\Leaf} | \theta_{\Leaf}) \, ,
\end{equation}
which is an exponentially large mixture efficiently represented by the hierarchical structure of the SPN.
For better distinguishability from product nodes which, as in classic SPNs, partition the output variables, we call product nodes that split a collection of input RVs at split points \emph{split nodes} throughout the rest of the paper.

An illustration of an SPN-GP with split nodes, product nodes and different types of kernel functions is shown in Figure~\ref{fig:SPN-P}.
As shown in the illustration, the SPN-GP model is a very flexible probabilistic regression model which allows to: (1) mix over different block-diagonal covariance representations (cf. $\SumNode_1$ in the illustration) (2) mix over GPs with different kernel functions (cf. $\SumNode_2$), (3) hierarchically sub-divides the input space (cf. the hierarchy of $\SplitNode_2, \SplitNode_3$ and $\SplitNode_2, \SplitNode_4$) and (4) partitions the dependent variables subject to the covariates (cf. $\ProductNode_2$ in the figure).
Further, this model allows accounting for structure in the input and the output space.

As an SPN-GP model is a deep structured mixture model over GP experts, the computation of the mean and variance for an unseen data point $\x\new$ is given by
\begin{align} \label{eq:posterior_SPN-GP}
  \mathbb{E}[f\new] &= \sum_{t=1}^\tau \prod_{(\SumNode, \Child) \in \SPT_{t, E}} &&\w_{\SumNode, \Child} \, \mathbb{E}[f_{\Leaf(\new, \SPT_{t})}] \, ,\\
  \mathbb{V}[f\new] &= \sum_{t=1}^\tau \prod_{(\SumNode, \Child) \in \SPT_{t, E}} &&\w_{\SumNode, \Child} ((\mathbb{E}[f_{\Leaf(\new, \SPT_{t})}] - \mathbb{E}[f\new])^2 \\ &&&+ \mathbb{V}[f_{\Leaf(\new, \SPT_{t})}]) \, ,
\end{align}
where $\Leaf(\new, \SPT_{t})$ selects the GP expert of the induced tree $\SPT_{t}$ responsible for $\x\new$. 

\subsection{Exact Posterior Inference}
In the Bayesian setting we wish to update the prior distribution over functions defined by an SPN-GP based on an observed dataset $\data = \{(\x_n,y_n)\}_{n=1}^N$.
Under the usual i.i.d.~assumption, the posterior can be written as
\begin{align}
 p(f \,|\, \mathcal{D}) \propto \prod^N_{n=1} p(y_n \,|\, f_n) \, p(f_n \,|\, \x_n). 
 \label{eq:SPNGP_unnormalized_posterior}
\end{align}
Here $p(f_n \,|\, \x_n)$ is an SPN-GP, i.e.~it can either be i) a mixture over SPN-GPs / GPs (sum nodes), ii) a product over SPN-GPs / GPs (product node), or iii) a GP (leaf).
In the first case we can write \eqref{eq:SPNGP_unnormalized_posterior} as
\begin{eqnarray}
  \begin{aligned}
 p_{\SumNode}(f \,|\, \mathcal{D}) 
 & \propto \prod_{n\in N_\SumNode} p(y_n \,|\, f_n) \sum_{\Child \in \ch(\SumNode)} \w_{\SumNode,\Child} \, p_{\Child}(f_n \,|\, \x_n)  \\
 & = \sum_{\Child \in \ch(\SumNode)} \w_{\SumNode,\Child} \, \prod_{n\in N_\SumNode} p(y_n \,|\, f_n) \, p_{\Child}(f_n \,|\, \x_n),
 \label{eq:SPNGP_unnormalized_posterior_sum}
 \end{aligned}
\end{eqnarray}
i.e.~we can ``pull'' the likelihood terms over the sum, where $N_\SumNode$ is the set of data point indices which are assigned to node $\SumNode$.
In case ii), we can write \eqref{eq:SPNGP_unnormalized_posterior} as
\begin{eqnarray}
 \begin{aligned}
 p_{\ProductNode}(f \,|\, \mathcal{D}) 
 & \propto \prod_{n\in N_\ProductNode} p(y_n \,|\, f_n) \prod_{\Child \in \ch(\ProductNode)} \, p_{\Child}(f_n \,|\, \x_n)  \\
 & = \prod_{\Child \in \ch(\ProductNode)} \left( \prod_{n \in N_{\Child}} p(y_n \,|\, f_n) \, p_{\Child}(f_n \,|\, \x_n)  \right),
 \label{eq:SPNGP_unnormalized_posterior_prod}
 \end{aligned}
\end{eqnarray}
where $\bigcup\limits_{\Child \in \ch(\ProductNode)} N_{\Child} = N_\ProductNode$ with $\bigcap\limits_{\Child \in \ch(\ProductNode)} N_{\Child} = \emptyset$.
Inductively repeating this argument for all internal nodes, we see that we obtain an unnormalized posterior by multiplying each GP leaf with its local likelihood terms.
We can therefore efficiently perform exact posterior updates in SPN-GPs by application of posterior inference at the leaves and re-normalization of the SPN \cite{Peharz2015}. 

\begin{figure*}%
    \centering
    \includegraphics[width=0.49\linewidth]{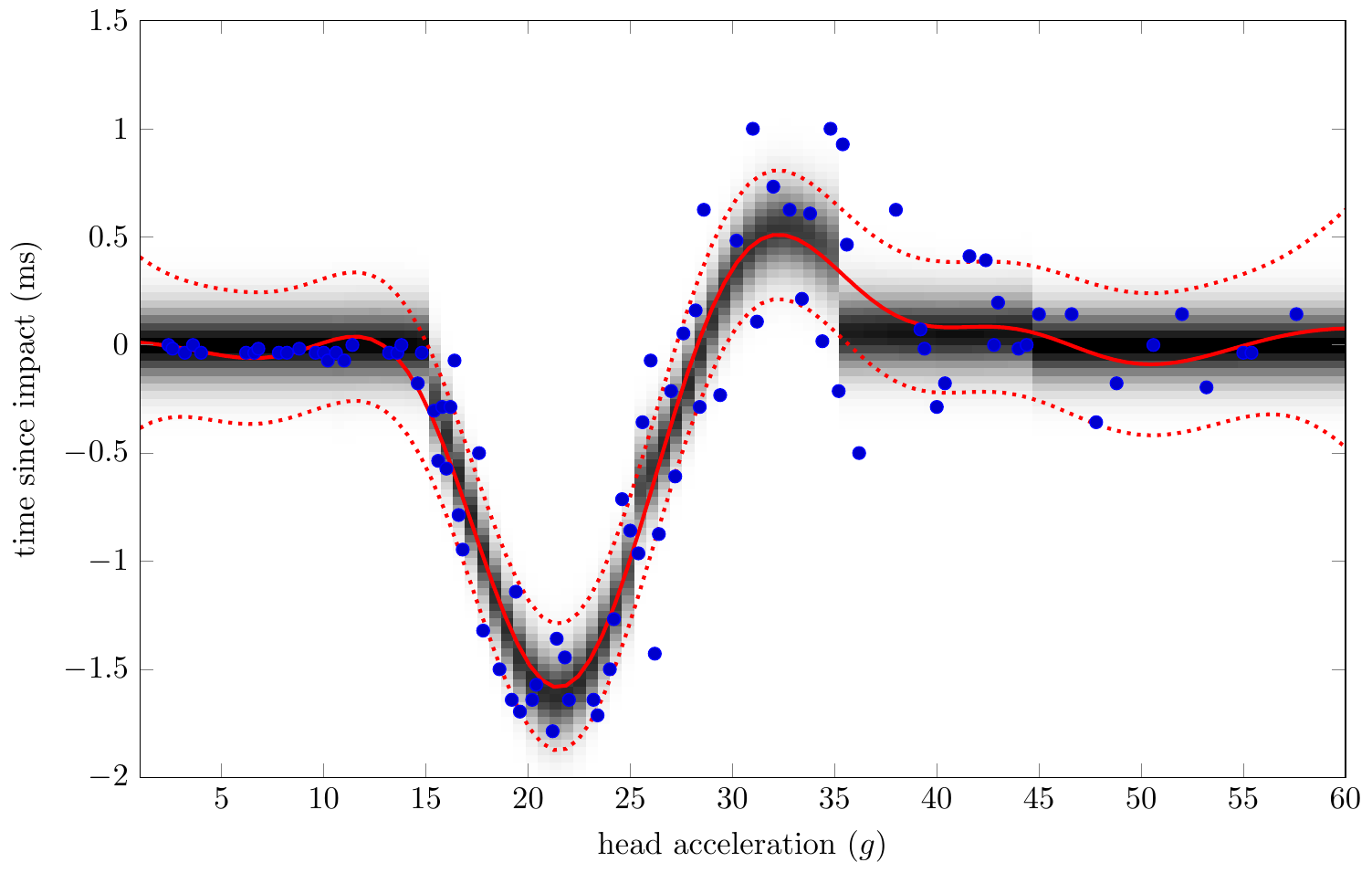}
    \includegraphics[width=0.49\linewidth]{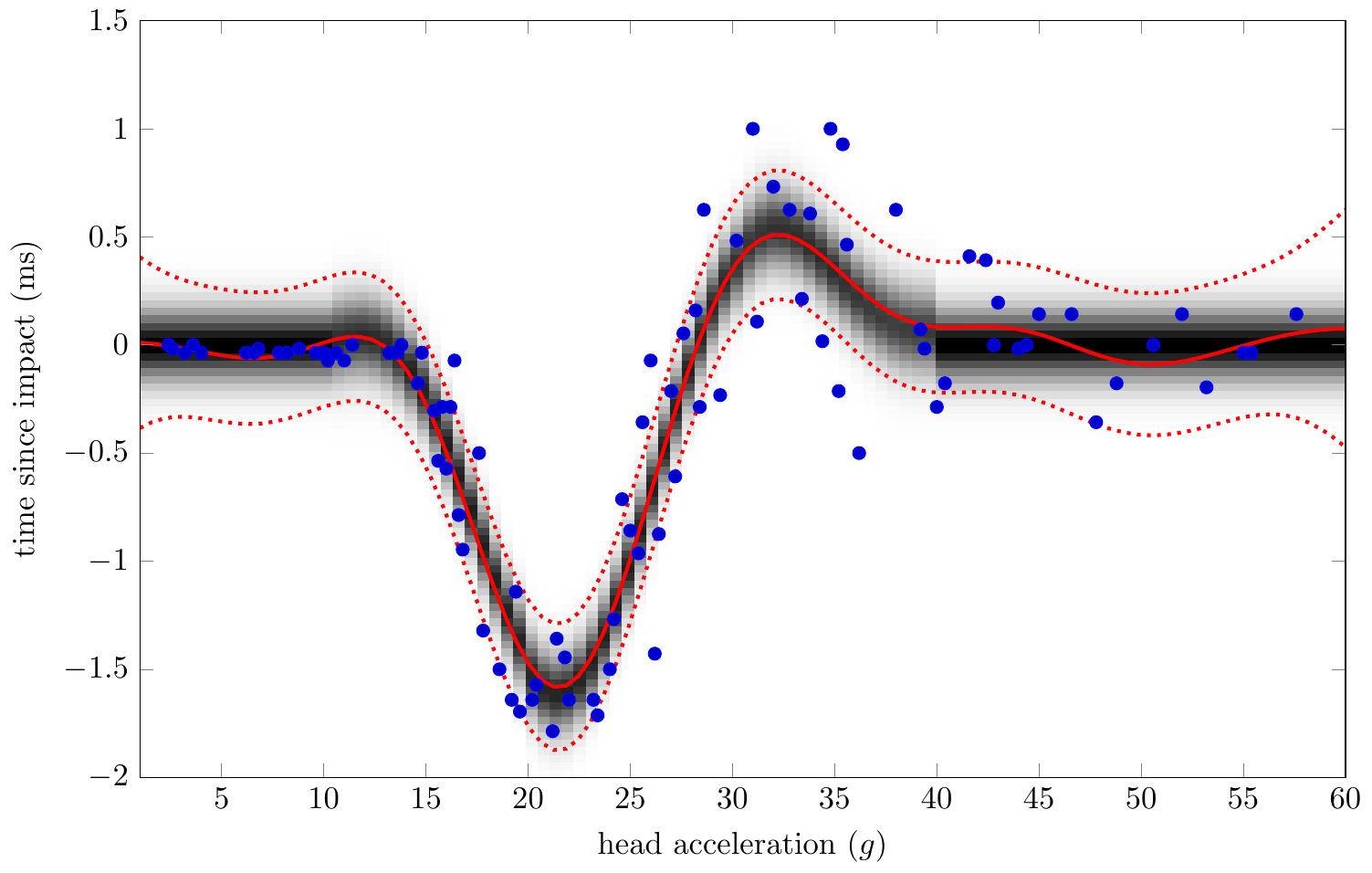}
    \caption{[Best viewed in colour] (left) Density region of an SPN-GP, coloured in grey, in comparison to a traditional GP, indicated in red (solid line shows the mean and the dotted line the 95\% confidence interval). As shown, the subdivision of the input space can result in discontinuities in the density region. (right) However, these discontinuities can be drastically reduced by allowing training points at region boundaries to flow across the boundary. Note that the SPN-GP is able to select optimal kernel functions depending on the input.}%
    \label{fig:motorcycle}%
\end{figure*}

\subsection{Hyper-parameter Optimization}
As for GPs, we can find optimal hyper-parameters by maximizing the marginal likelihood of an SPN-GP.
Due to decomposability of valid SPN-GPs, cf. previous section, we can pull down the integration over latent functions to the leaf nodes of the network, i.e. 
\begin{equation}
    \begin{aligned}
    p(\y | \X, \theta) &= \bigintsss_f \sum_{t=1}^\tau p(\SPT_{t}) \prod_{\Leaf \in \SPT_{t, V}} p(\bm \y_{\Leaf} | f_\Leaf, \X_\Leaf) p(f_\Leaf |\X_\Leaf) df \, , \\
   &= \sum_{t=1}^\tau p(\SPT_{t}) \prod_{\Leaf \in \SPT_{t, V}} p(\bm \y_{\Leaf} | \X_\Leaf)\, ,
   \end{aligned}
\end{equation}
where $p(\SPT_{t}) = \prod_{(\SumNode, \Child) \in \SPT_{t, E}} \w_{\SumNode, \Child}$ and $(\X_{\Leaf}, \bm \y_{\Leaf})$ are the observed points and function values at leaf node $\Leaf$.
Therefore, to find appropriate hyper-parameters for the GP experts in an SPN-GP, we can maximize the marginal likelihood of each expert independently or maximize the marginal likelihood of the SPN-GP jointly, if hyper-parameters are tied.

\subsection{Structure Learning}
To learn the structure of SPN-GPs we extend the approach introduced by Poon and Domingos \cite{Poon2011} to construct network structures for different regression scenarios.
Our algorithm first constructs a region graph \cite{Poon2011,Dennis2012} which hierarchically splits the input space into small regions.
The recursive subdivision stops if a constructed region contains less than $O$ many points or if a region cannot be split into sub-regions anymore.
Our approach constructs axis aligned partitions at each level of the recursion with the dimension in question being randomly select.
Starting from the region graph, each leaf region is equipped with multiple experts, each of which has an individual kernel function.
These experts are combined in the network structure and combinations of these experts, i.e. induced trees, are weighted according to their plausibility. 
Therefore, the proposed model naturally is a weighted mixture over local experts with each component having multiple local experts with different kernel functions.
We refer to the appendix for further details on the structure learning algorithm.
Note that for uniformly distributed observation and equally sized sub-regions under split node, SPN-GPs constructed with our structure learning algorithm reduce the computation complexity of GPs, i.e. $\mathcal{O}(N^3)$, to $\mathcal{O}(S*K \frac{N^3}{2^D})$ where $S$ is the number of splitting schemes ($S = 1$ in all experiments), $K$ is the number of kernel functions and $D$ is the depth of the network, i.e. the number of consecutive split nodes.

\section{Experiments}   \label{sec:experiments}
To evaluate the effectiveness and investigate the properties of SPN-GPs for probabilistic regression, we conducted a series of qualitative and quantitative evaluations.
The first two datasets, i.e. the motorcycle and the synthetic dataset, are used to analyze the capacities and challenges arising in the SPN-GP model.
While we used three UCI datasets from the UCI machine learning repository \cite{UCI} to quantitatively evaluate the performance of SPN-GPs against linear models and traditional GPs and additionally conducted an evaluation using the Kin40K dataset to assess the performance of SPN-GPs against other models which leverage hierarchies of local experts. 

\subsection{Qualitative Evaluation} \label{sec:qualitative}
To analyze the capacities and the challenges arising in SPN-GPs, we qualitatively evaluated the SPN-GP model on the motorcycle dataset \cite{Silverman1985} and on a synthetically generated time series with non-stationarities.
The synthetic dataset was generated using the function \texttt{demo\_epinf} described in \cite{Vanhatalo2013} and has been used in prior work to asses the performance of GP models for non-stationary series, e.g. \cite{Heinonen2016}.
We used a minimum sub-region size of $\delta = 10$, linear and squared exponential kernel functions for the motorcycle dataset and linear and Mat\'{e}rn kernel functions for the synthetic dataset. 
In both cases we optimized hyper-parameters while for the motorcycle dataset we fixed the noise variance to $\sigma_{\epsilon} = 0.37$. 
For comparison, we used a traditional GP with a squared exponential kernel function and optimized the hyper-parameters as in the case of the GP-SPN by maximizing the marginal log-likelihood.

Figure~\ref{fig:motorcycle} shows the density region of the SPN-GP in grey and illustrates the GP in red (mean and 95\% confidence interval).
We can see that the independence assumptions made be the SPN-GP result in discontinuities in the density region.
In particular, the boundary around position $35$ results in strong discontinuities due to the enforced independence assumption and the change of kernel function.
However, as illustrated in Figure~\ref{fig:motorcycle} (right-hand side) those discontinuities can be drastically reduced by accounting for outlying data points close to a regional boundary during posterior inference at the leaves.
This approach lets us account for dependencies of points spatially close to the boundary.
Note that care has to be taken to ensure the validity of the SPN-GP, e.g. updates of the weights have to exclude the additional boundary points. 
The approach is based on a similar approach recently used in a hierarchical model \cite{Ng2014}.
We can observe that in both cases, the SPN-GP is able to weight the different kernel functions according to their plausibility for each region independently. 
In particular, in regions with mostly linear dependencies, the linear kernel function was selected as the most plausible kernel, while in other regions the network prefers to use a squared exponential kernel function.

\begin{figure*}%
    \centering
    \includegraphics[width=0.49\linewidth]{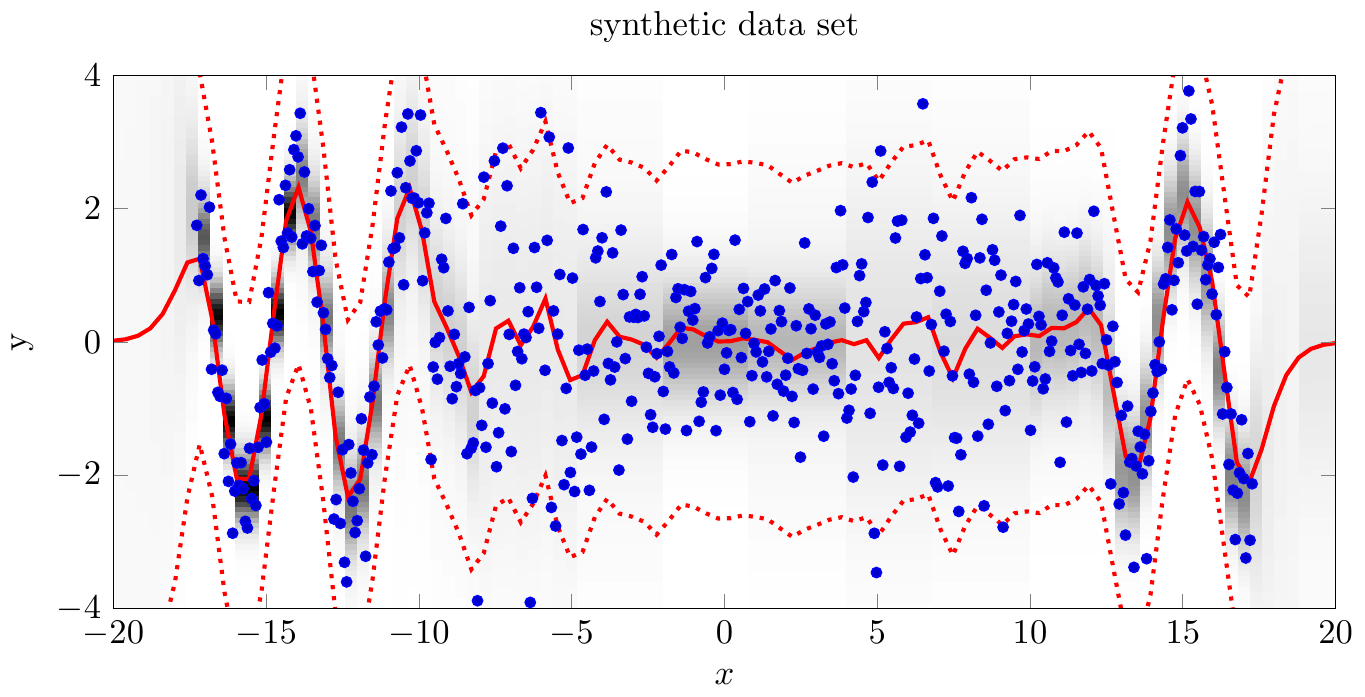}
    \includegraphics[width=0.49\linewidth]{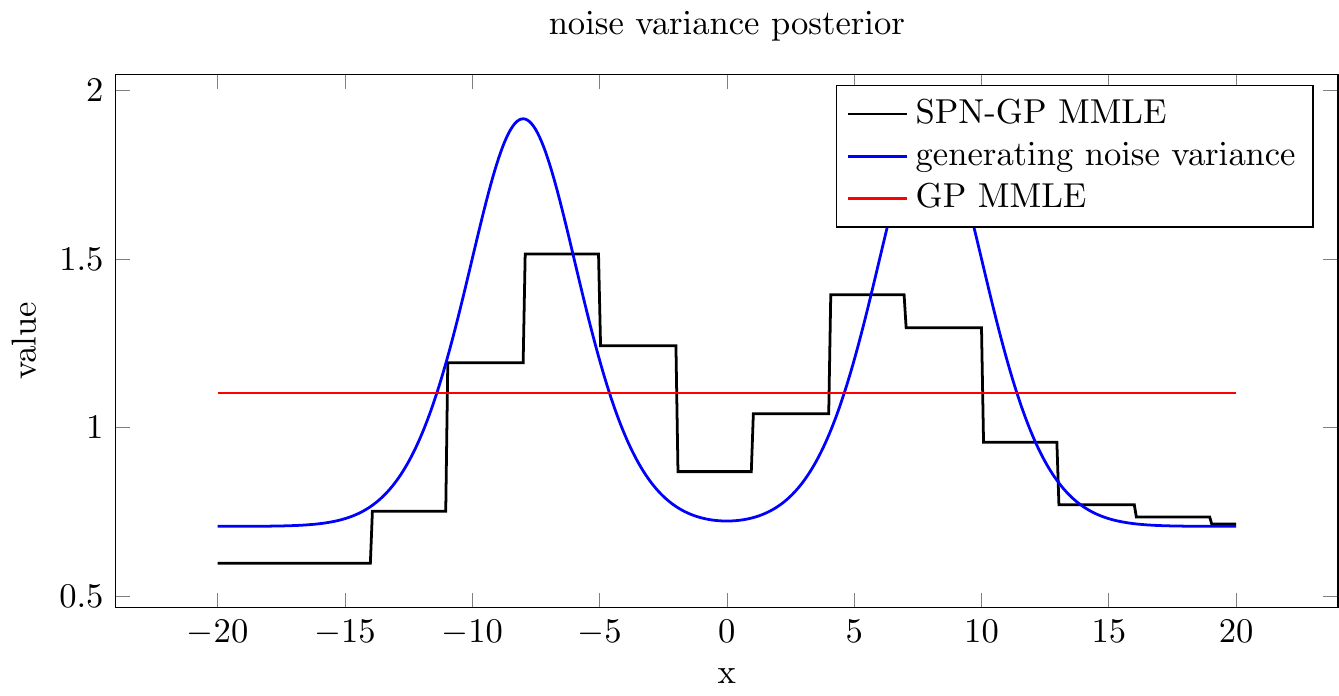}
    \caption{[Best viewed in colour] (left) Density region of an SPN-GP, coloured in grey, compared to a traditional GP, indicated using red lines. 
    The SPN-GP is capable of adapting the noise variance, the kernel hyper-parameters and weights the kernel functions depending on the observations.
    (right) Comparison of the maximum marginal likelihood estimates (MMLE) for the noise variance of an SPN-GP (black line) and a traditional GP (red line). The true noise generating function is shown in blue.}%
    \label{fig:nonstationary}%
\end{figure*}

Figure~\ref{fig:nonstationary} shows the density region of the SPN-GP and the mean and 95\% confidence interval of a GP for non-stationary data with conditional heteroscedastic noise.
Even for more complex functions, the SPN-GP is able to match the general shape of the traditional GP.
Further, we can see on Figure~\ref{fig:nonstationary} (right-hand side) that the SPN-GP model is able to adjust the noise variance parameter depending on the input (black line) approximately matching the true noise generating function (blue line).
Even though the SPN-GP model is not explicitly designed to work for non-stationary time series, the subdivision of the input space allows the model to naturally account for input dependent hyper-parameters.
We believe that this is an exciting side effect of the SPN-GP model and illustrates the flexibility of the model. 

\subsection{Quantitative Evaluation}
To quantitatively evaluate SPN-GPs we computed the root mean squared error (RMSE) on different UCI datasets and on the Kin40k dataset.
On the UCI datasets, we assessed the performance of the SPN-GP model against the mean prediction (Mean), linear least squares (LLS), ridge regression (Ridge) with $\alpha = 0.01$ and a Gaussian process (GP) for those datasets where the computation is feasible on a MacBook Pro with 8 GB RAM.
For the GP model, we used the best performing kernel function (out of linear, Mat\'{e}rn and squared exponential kernel) with fixed hyper-parameters. 
We tried to optimize the hyper-parameters of the GP but obtained worse results than using hand-picked parameters.
We suspect that the decrease in performance is due to overfitting of the GP on the training examples.
Therefore, we did not optimize the hyper-parameters of the GP and the SPN-GP model for fair comparisons.
Note that the hyper-parameters have been selected in favour of the traditional GP model.
We equipped the SPN-GP model with linear, Mat\'{e}rn and squared exponential kernel functions and learned the structure with $O = 500$, $S = 1$ and a recursive subdivision of the input space into four equally sized sub-regions.
We additionally evaluated the effect of overlapping, as discussed in Section~\ref{sec:qualitative} denoted as SPN-GP$^*$.

The resulting RMSE and standard errors computed over five independent runs are listed in Table~\ref{tab:uci_rmse}.
On the \textsc{Concrete} dataset the SPN-GP model achieves competitive results compared to a GP regressor.
While on the \textsc{Energy} dataset our model outperforms traditional GPs when overlapping regions are used.
Further, we found a positive effect of reducing the discontinuities, cf. SPN-GP$^*$ in Table~\ref{tab:uci_rmse}, in two out of three datasets indicating that the performance of the SPN-GP sometimes suffers from the independence assumptions made in the model architecture.

\begin{table}
\caption{Root mean squared errors (RMSE) and standard errors of the SPN-GP model compared to mean prediction (Mean), linear least squares (LLS), ridge regression (Ridge) and Gaussian process regression (GP). SPNs with additional overlap are denoted as SPN-GP$^*$. Smaller is better.}
\label{tab:uci_rmse}
\vskip 0.15in
\begin{center}
\begin{small}
\begin{sc}
\begin{tabular}{l|ccc}
\toprule
Method & Energy & Concrete & CCPP\\
\midrule
Mean & $9.83 \pm 0.09$ & $16.45 \pm 0.10$ & $17.00 \pm 0.05$ \\
LLS & $3.08 \pm 0.02$  & $10.33 \pm 0.25$ & $4.63 \pm 0.04$  \\ 
Ridge & $3.08 \pm 0.02$ & $10.33 \pm 0.25$ & $4.63 \pm 0.04$  \\
GP & $2.44 \pm 0.17$ & $\bm{6.25} \pm 0.14$ & -  \\
SPN-GP & $2.23 \pm 0.11$ & $6.27 \pm 0.20$ & $\bm{4.10} \pm 0.05$  \\
SPN-GP$^*$ & $\bm{2.07} \pm 0.04$ & $\bm{6.25} \pm 0.14$ & $4.11 \pm 0.04$  \\
\bottomrule
\end{tabular}
\end{sc}
\end{small}
\end{center}
\vskip -0.1in
\end{table}

\begin{figure*}%
    \centering
    \includegraphics[width=\linewidth]{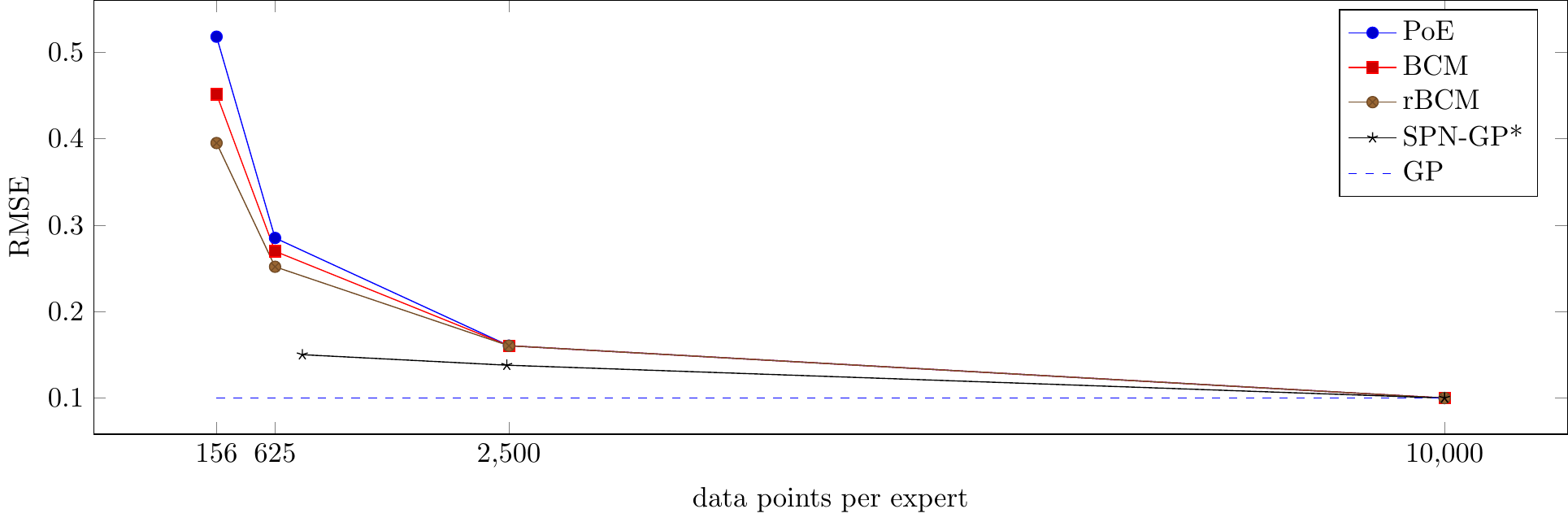}
    \caption{[Best viewed in colour] Approximation error of SPN-GPs compared to Product of Experts (PoE), Bayesian Committee Machine (BCM) and the robust Bayesian Committee Machine (rBCM) as a function of number of points per expert. Ground truth RMSE of a full GP is shown as a dashed line.}%
    \label{fig:kin40k}%
\end{figure*}

Additionally, we assessed the approximation error on the Kin40K dataset as described in \cite{Deisenroth2015}.
The Kin40K dataset consists of $10,000$ training examples and $30,000$ test points where each data point represents the forward dynamics of an 8-link all-revolute robot arm.
The dependent variable in this dataset is the distance of the end-effector from a target.
We followed the approach described in \cite{Deisenroth2015} and used hyper-parameters of a squared exponential kernel (with ARD) estimated using a full GP and compared the RMSE against different numbers of data points per expert.
As our structure learning algorithm does not guarantee to construct sub-regions with an equal number of training points, we depict the average number of training points per expert for SPN-GPs.
The performance of SPN-GPs compared to state of the art approaches is shown in Figure~\ref{fig:kin40k}.
Note that our model architecture used in this experiment is comparable to prior work, i.e. only a single kernel function but mixing over split positions is used, and is not more expressive than a full GP.
As shown in Figure~\ref{fig:kin40k} the RMSE of SPN-GPs rises only slightly with declining number of training points per expert indicating that our approach does not suffer from vanishing predictive variances or the effect of weak experts.

\section{Discussion and Conclusion}   \label{sec:discussion}
In this work, we have introduced sum-product networks with Gaussian process leaves (SPN-GPs) as an efficient and effective probabilistic non-linear regression model.
In our model, the inference cost is effectively reduced in a divide and conquer fashion by recursively sub-dividing the input space into small sub-regions and distributing the inference onto local experts. 
We showed that SPN-GPs have a variety of appealing properties, such as 
(1) computational and memory efficiency, 
(2) efficient and exact posterior inference, 
(3) the ability to weight different kernel functions according to their plausibility, and
(4) naturally account for non-stationarities in data.

We found that the SPN-GP model is on par with or outperforms traditional GP models.
Moreover, we could demonstrate that our model does not suffer from vanishing predictive variances or the effect of weak experts.
In addition, we assessed the properties and challenges of our model qualitatively.
We found that SPN-GPs naturally account for non-stationarities in time series and that SPN-GPs are able to effectively infer the appropriate kernel function depending on the input.
Our experiments indicate that the SPN-GP model is a very promising model for non-linear probabilistic regression not only in large-scale data domains.
Further, we believe that this work will lead to interesting new approaches for regression in high-dimensional data regimes, e.g. hyper-parameter optimization.
In future work, we will investigate the application of SPN-GPs for large-scale real-world problems and work on data-driven approaches for structure learning.
Further, we plan to investigate the effects of parameter tying for regularization in SPN-GP models.

\subsubsection*{Acknowledgments}
This research is partially funded by the Austrian Science Fund (FWF): P 27803-N15.

\bibliographystyle{icml2018}
\bibliography{references}

%\pagebreak
\newpage
\appendix
\section{Structure Learning}
To learn the structure of SPN-GPs we extended the approach by Poon and Domingos \cite{Poon2011} to construct network structures for different regression scenarios.
Our Algorithm~\ref{alg:structLearning} first constructs a region graph \cite{Poon2011,Dennis2012} by using Algorithm~\ref{alg:structLearning_region_graph} which hierarchically splits the input space into small regions.
In particular, our algorithm starts from the root region and selects an axis at random. This axis is then used to partition the input space into disjoint sub-regions.
Subsequentially, each sub-region is recursivelly processed in the same way as the root region.
The recursive subdivision stops if a constructed region contains less the $O$ many points or if a region cannot be split into sub-regions.

After constructing the region graph Algorithm~\ref{alg:structLearning} equips each leaf region with a set of GP experts, each using a different kernel function. 
And equips the partitions with split nodes and the internal regions with sum nodes resepectivelly.
Note that it is possible to mix over different partitions of the input space by constructing partitions with different offsets under the root region.
However, this approach can lead to an increase of computational complexity and the construction therefore has to be done carefully. 

\begin{algorithm}[tb]
   \caption{Region Graph Construction}
   \label{alg:structLearning_region_graph}
\begin{algorithmic}[1]
   \STATE {\bfseries Input:} $\data = \{(\xn, \y_n)\}_{n=1}^{N}$ (observations), $\delta = \{\delta_d\}_{d=1}^D$ (subinterval sizes), $O$ (minimum number of samples).
   \STATE $\mathcal{R} \leftarrow$ new empty region graph
   \STATE $R \leftarrow$ new empty region
   \STATE $Q \leftarrow$ new Queue
   \STATE insert $R$ into $Q$ and $\mathcal{R}$
   \WHILE {$Q$ is not empty}
   \STATE $R \leftarrow$ pop region from $Q$
   \IF {number of samples in $R$ larger then $O$}
   \STATE $d \leftarrow$ random($1, \dots, D$)
   \STATE $\bm P \leftarrow$ new partitions of $R$ in $d$ with subintervals of size $\delta_d$
   \FORALL{$P \in \bm P$}
    \STATE connect $R \rightarrow P$ and insert $P$ into $\mathcal{R}$
    \STATE $\bm R \leftarrow$ split $R$ in dimension $d$ on position $s_P$
    \FOR{$\hat{R} \in \bm R$}
    \IF{$\hat{R} \notin \mathcal{R}$}
    \STATE insert $\hat{R}$ into $Q$ and $\mathcal{R}$
    \ENDIF
    \STATE connect $P \rightarrow \hat{R}$
    \ENDFOR
    \ENDFOR
    \ENDIF
   \ENDWHILE
   \STATE {\bfseries Return:} Region graph $\mathcal{R}$
\end{algorithmic}
\end{algorithm}

\begin{algorithm}[tb]
   \caption{SPN-GP Structure Construction}
   \label{alg:structLearning}
\begin{algorithmic}[1]
   \STATE {\bfseries Input:} $\data = \{(\xn, \y_n)\}_{n=1}^{N}$ (observations), $\delta = \{\delta_d\}_{d=1}^D$ (subinterval sizes), $O$ (minimum number of samples), $S$ (number of sum nodes), $\bm k \in \{k_l\}_{l=1}^L$ (kernel functions).
   \STATE $\mathcal{R} \leftarrow $ construct new region graph using Algorithm~\ref{alg:structLearning_region_graph}
    \STATE $\SPN \leftarrow$ empty SPN-GP
    \FOR {$R \in \mathcal{R}$}
        \IF {$R$ has no child}
            \FOR {$k \in \bm k$}
                \STATE equip $R$ with GP leaf $\Leaf$ using kernel function $k$
                \STATE condition GP on $(\x, \y) \in R$
            \ENDFOR
        \ELSIF {$R$ has no parent}
            \STATE equip $R$ with a sum node $\SumNode$
        \ELSE
            \STATE equip $R$ with $S$ sum nodes $\bm \SumNode$
        \ENDIF
    \ENDFOR
    \FOR {$P \in \mathcal{R}$}
        \STATE $R_1, R_2 \leftarrow$ sub-regions of $P$
        \FOR{$\SumNode_1 \in R_1$, $\SumNode_2 \in R_2$}
            \STATE equip $P$ with split node $\ProductNode = \SumNode_1 \times \SumNode_2$
            \STATE $\hat{R} \leftarrow$ find $(R_1 \cup R_2)$ in $\mathcal{R}$
            \FORALL{$\SumNode \in \hat{R}$}
                \STATE connect $\SumNode \rightarrow \ProductNode$
            \ENDFOR
        \ENDFOR
    \ENDFOR
   \STATE {\bfseries Return:} SPN-GP $\SPN$
\end{algorithmic}
\end{algorithm}

\section{Quantiative Evaluation}
To quantitatively evaluate SPN-GPs we evaluated the performance of our model on the UCI datasets listed in Table~\ref{tab:uci}.

\begin{table}[h]
\caption{Statistics of UCI datasets used for quantitative evaluation. The number of sample is depicted by $N$, the number of input dimensions using $D_X$ and $D_Y$ denotes the number of dependent variables.}
\label{tab:uci}
\vskip 0.15in
\begin{center}
\begin{small}
\begin{sc}
\begin{tabular}{l|ccc}
\toprule
Data set & $N$ & $D_X$ & $D_Y$ \\
\midrule
Energy & 768 & 8 & 2 \\
Concrete & 1030 & 8 & 1 \\
CCPP & 9568 & 4 & 1 \\
\bottomrule
\end{tabular}
\end{sc}
\end{small}
\end{center}
\vskip -0.1in
\end{table}

\end{document}